\documentclass[11pt,a4paper]{article}
\usepackage[T1]{fontenc}
\usepackage[utf8]{inputenc}
\usepackage[margin=25mm]{geometry}
\usepackage{mathptmx}
\usepackage{amsmath,amssymb,graphicx}
\usepackage{booktabs,array,tabularx}
\usepackage[authoryear,round]{natbib}
\usepackage{microtype,setspace,enumitem,caption,xcolor}
\usepackage[hidelinks,breaklinks]{hyperref}
\usepackage{url}
\usepackage{flafter}
\usepackage[section]{placeins}
\newcommand{\doi}[1]{\url{https://doi.org/#1}}
\setlist{nosep,leftmargin=*}
\newcolumntype{L}[1]{>{\raggedright\arraybackslash}p{#1}}
\newcolumntype{Y}{>{\raggedright\arraybackslash}X}
\newcommand{\E}{\mathbb{E}}

\newcommand{\I}{\mathcal{I}}
\newcommand{\F}{\mathcal{F}}

\hypersetup{pdftitle={When does a scaling result justify a different allocation? A critical review of resource-allocation evidence for AI systems},pdfauthor={Seyed Emadi}}

\title{\LARGE\bfseries When does a scaling result justify a different allocation?\\A critical review of resource-allocation evidence for AI systems}
\author{Seyed Emadi\\[3pt]
\normalsize Kenan-Flagler Business School\\
\normalsize University of North Carolina at Chapel Hill, Chapel Hill, NC, USA\\
\normalsize Correspondence: \texttt{seyed\_emadi@kenan-flagler.unc.edu}\\
\normalsize ORCID: 0000-0002-4571-730X}
\date{}

\begin{document}
\maketitle
\begin{abstract}
AI scaling studies increasingly evaluate systems that combine a pretrained model with retrieval, search, verification, tools, and interaction. Yet a higher score under a larger budget does not by itself show where additional resources are best spent. This critical integrative review asks when a reported scaling result supports a resource-allocation decision. It compares evidence across pretraining, test-time computation, retrieval, and agent evaluation, distinguishing the performance of a tested procedure from the best performance achievable under a resource limit. The synthesis shows that three mismatches recur across this evidence: success counted before an answer is chosen, information a deployed system will not have, and costs left out of the comparison. A capability surface expresses performance as a function of budgets, mechanisms, and available information. Worked analytical examples show how the evaluation metric, deployment volume, selection rule, and stopping policy can alter an allocation conclusion. A resource envelope provides a structured record of the task, development and run-time resources, information access, and procedure behind a reported score. Its application to a published comparison illustrates which conclusions the evidence supports and which deployment questions remain unresolved. The resulting framework specifies the comparisons needed to choose among feasible systems and motivates experiments on the transfer of allocation rules across tasks and operating conditions. It does not propose a universal scaling law or infer general intelligence from benchmark gains.
\end{abstract}

\section{Introduction}
\label{sec:intro}
The practical success of scaling laws comes from their conditional precision. Within a specified model family, data distribution, and training procedure, researchers can estimate how predictive loss changes with parameters, processed tokens, and computation \citep{kaplan2020scaling,hoffmann2022training}. This makes expensive development decisions more tractable. However, a deployed artificial intelligence (AI) system increasingly makes additional decisions: whether to retrieve a document, invoke a tool, generate another candidate, consult a verifier, or stop. These decisions determine both the answer and the resources consumed. A parameter count cannot describe all of them.

This review asks: \emph{When does a reported scaling result justify allocating resources differently?} Three terms need defining before it can be answered.

\noindent\textbf{The allocation problem.} A team has a budget, in money or compute, to build and run an AI system for a task. It can spend the budget in two places. It can spend on building: a larger model, more training data, a trained verifier, a retrieval index. It can spend on running: for each query, more candidate answers, a longer search, retrieval calls, verifier calls. The allocation problem is to divide the budget between and within these two places so that the system meets the task's quality requirement at the lowest total cost, using only the information it will have in deployment.

\noindent\textbf{A scaling result.} A finding about one piece of this problem: how quality responds to one resource, with everything else held at one setting.

\noindent\textbf{An allocation claim.} The assertion that one division of the budget is better than another. A scaling result identifies an allocation claim when it establishes that claim.

Consider a question-answering service built on one language model. Its team must decide whether to buy a larger model or to keep the current model and run it many times per question, picking the best of its answers. A published study reports that the second option scores higher for the same compute. Does that settle the team's decision? Not yet. The study's score was produced under conditions the team may not share. Three are common.

First, how success was counted. Many studies generate, say, 100 answers per question and count the question as solved if any one of the 100 is correct. Studies call this measure coverage. The team's users receive one answer. Unless the system can pick the correct answer out of the 100, the users see none of that success. The share of questions on which the one returned answer is correct is the returned-answer accuracy, and that is the number the team needs.

Second, what the system was allowed to know. Some studies use the benchmark's answer key during the experiment, for example to sort questions by difficulty before deciding how much computation each gets, or to pick the returned answer. In production there is no answer key.

Third, what was counted as cost. The study may measure cost in floating-point operations (FLOPs) and count only the model's generation. The team pays in dollars and latency, and it also pays to train the verifier that does the picking, a cost the study may leave out.

In each case the study is right about what it measured. The team's decision depends on things it did not measure.

The review makes three contributions. First, it sorts the scaling literature by what each study can tell a team. A study may show that a procedure improves with more of a resource, or why it improves. It may show that the procedure beats the alternatives at the same cost, or say something about alternatives that were not tested. Second, it shows that the three mismatches just described recur across the literature: success counted before an answer is chosen, information a deployed system will not have, and costs left out. Third, it proposes a record, the resource envelope, in which a study states those three things alongside its score, so that a reader can see at once whether any mismatch applies. It is a model card for a result, with the costs and the information the system had attached. Together, these give a reader a way to tell what an experiment established and what remains to be shown before acting on it.

The underlying ideas have substantial precedents. Rational metareasoning treats computation as an action whose expected benefit must justify its cost \citep{russell1991metareasoning}, and anytime algorithms relate solution quality to available deliberation time \citep{zilberstein1996anytime}. More recent work asks the same question of one component at a time. It asks how to split compute between training and inference, how to choose the models and calls in a pipeline, and how to judge an agent by cost as well as accuracy. This review builds on all of it.

The discussion focuses on large language model (LLM) systems because this is where the reviewed evidence is concentrated. Three kinds of task recur in it: producing a result that must then be checked, predicting what an intervention would do, and carrying out a long chain of steps without losing track. The review uses them as test cases for allocation claims. They are not offered as a definition of intelligence; definitions and measures of intelligence remain diverse \citep{legg2007universal,chollet2019measure,morris2024levels,hernandezorallo2017measure}. The question here is narrower: which resource changes improve a given capability, at what cost, and under what conditions.

\begin{figure}[!htbp]
\centering
\includegraphics[width=\textwidth]{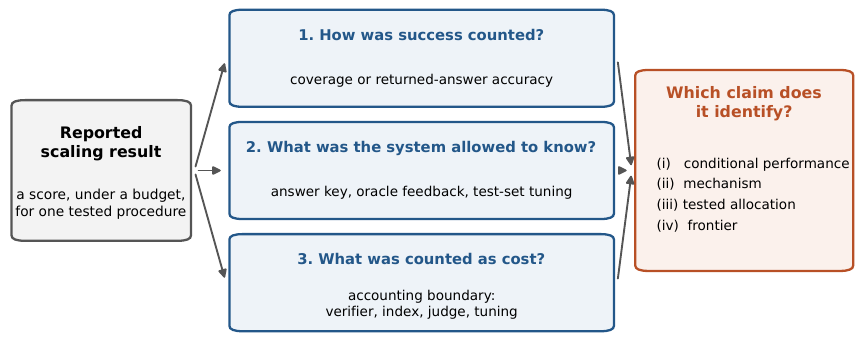}
\caption{\textbf{Reading a scaling result as evidence for an allocation decision.} A reported score is read through three questions: how success was counted, what the system was allowed to know, and what was counted as cost. The answers determine which of four kinds of claim the result identifies. The resource envelope of Section~\ref{sec:envelope} records the answers with the score, and a missing answer names the next experiment.}
\label{fig:reading}
\end{figure}

Section~\ref{sec:approach} explains how the studies were chosen and what earlier work already covers. Section~\ref{sec:framework} sets out the framework the review uses to read a study, and Section~\ref{sec:synthesis} applies it to the scaling literature study by study. Section~\ref{sec:bottlenecks} turns to the three kinds of task, asking what limits a system on each and therefore which resource to test next. Section~\ref{sec:examples} works through numerical examples in which the metric, the selection rule, the deployment volume, or the stopping rule changes the answer. Section~\ref{sec:envelope} defines the resource envelope and fills it in for one published study. Section~\ref{sec:agenda} proposes experiments; Section~\ref{sec:conclusion} states limitations and concludes. Figure~\ref{fig:reading} shows how these parts fit together.

\section{Review approach and relation to existing syntheses}
\label{sec:approach}
An allocation claim needs three facts from a study: what was varied, what was held fixed, and what was measured. The studies in this review were chosen because they state those facts, or because they show what goes wrong when the facts are missing. This section first explains how the studies were gathered and read, then relates the review to earlier work that asked the same question. Together these set the scope for the framework in Section~\ref{sec:framework}.

\subsection{Scope, search, and evidence handling}
This is a critical review, not a systematic review or a meta-analysis: it does not count studies or pool their results. The search closed on 9 September 2026. It began from the foundational scaling and evaluation papers and followed their citations in both directions. Targeted searches then covered scaling laws, the split of compute between training and inference, test-time computation, retrieval and datastore scaling, verification and repeated sampling, agent coordination, metareasoning, and resource-aware evaluation. Narrower searches covered distillation, quantization, speculative decoding, and serving. Sources came from arXiv, conference proceedings, OpenReview, the ACL Anthology, and publisher records. A study was included if it varied a resource or mechanism, analyzed a computational or information constraint, or stated an evaluation principle needed to interpret such variation. Methods papers were included where a mechanism needed explaining. Compression and serving methods appear only for their effect on what a budget can buy. Application surveys with no transferable argument about allocation were left out. Table~\ref{tab:criteria} summarizes the scope.

\begin{table}[!htbp]
\centering\small
\caption{Scope of the review: search window, sources, and inclusion criteria.}
\label{tab:criteria}
\begin{tabularx}{\textwidth}{@{}L{3.4cm}Y@{}}
\toprule
Element & This review\\
\midrule
Search window & Closed 9 September 2026.\\[3pt]
Sources & arXiv, conference proceedings, OpenReview, the ACL Anthology, and publisher records.\\[3pt]
Search method & Foundational scaling and evaluation papers, their citations followed in both directions, and targeted searches on scaling laws, the training--inference split, test-time computation, retrieval and datastore scaling, verification and repeated sampling, agent coordination, metareasoning, and resource-aware evaluation. Narrower searches on distillation, quantization, speculative decoding, and serving.\\[3pt]
Included & A study that varied a resource or mechanism, analyzed a computational or information constraint, or stated an evaluation principle needed to interpret such variation.\\[3pt]
Included with a limited role & Methods papers where a mechanism needed explaining. Compression and serving methods, for their effect on what a budget can buy.\\[3pt]
Excluded & Application surveys with no transferable argument about allocation.\\[3pt]
Evidence handling & Peer-reviewed work anchors the synthesis. Preprints are marked in the bibliography and treated as provisional. The cited edition was read where methods, assumptions, or accounting mattered.\\
\bottomrule
\end{tabularx}
\end{table}
\FloatBarrier

For each study in the evidence tables, the review records what was varied, what was measured, and what was held fixed or left out, since that is what limits the allocation conclusion. Where that judgment depended on methods, assumptions, or cost accounting, the relevant passages of the cited edition were read, since these can differ between editions. Publication records were used to check bibliographic details, with AI assistance as disclosed in the declarations. No registered protocol or formal risk-of-bias instrument was used, so the tables are a chosen sample, not a census, and no counts or pooled effects are reported.

Peer-reviewed work anchors the synthesis. Recent preprints extend it, are marked as such in the bibliography, and are treated as provisional.

\subsection{What existing work already contributes}
Prior literature supplies several parts of the allocation problem. Metareasoning formalizes the value of further computation, and recent language-model work learns when intermediate reasoning is worthwhile \citep{russell1991metareasoning,desabbata2024rational}. Training--inference scaling incorporates serving demand into model development \citep{sardana2024beyond}. FrugalGPT and work on compound systems, pipelines that chain several models and calls, choose which model or call to use at each step \citep{chen2023frugalgpt,chen2025selector}. Agent evaluation already argues for joint optimization, cost control, and appropriate holdouts \citep{kapoor2024agents}.

Existing surveys also overlap substantially with the present scope. \citet{ke2025frontiers} organize reasoning by inference versus learning regimes and standalone versus agentic architectures. \citet{zhang2025ttssurvey} organize test-time scaling around what, how, where, and how well to scale, distinguishing candidate coverage from aggregation quality and discussing efficiency measures. \citet{chung2025ttssurvey} distinguish sampling, search, and trajectory optimization and examine the role of generative diversity. Verification design is itself the subject of a dedicated survey by \citet{venktesh2025verification}, including verifier types, supervision, and outcome versus process checking. This review takes the taxonomy of inference mechanisms and the importance of verification and efficiency from these surveys as given. What it adds is the allocation question, asked of each result.

\citet{niu2026testtime} connect test-time adaptation, learning, and scaling through feedback, including resource allocation and verification. The survey by \citet{liu2026reward} in this journal examines reward models for reasoning. Building on these syntheses, this review asks the narrower question of when reported evidence is sufficient to choose between two ways of spending the same budget. Its incremental contribution is to ask the same three questions of each comparison: how success was counted, what the system was allowed to know, and what was counted as cost. Table~\ref{tab:position} specifies that relationship without implying that preceding surveys ignore cost or systems.

\begin{table}[!htbp]
\centering\small
\caption{Relationship to prior work. This review adds a way of reading each result, three worked mechanisms, and a reporting record. It does not originate resource-aware AI.}
\label{tab:position}
\begin{tabularx}{\textwidth}{@{}L{3.5cm}YY@{}}
\toprule
Prior contribution & Established focus & What this review adds\\
\midrule
Metareasoning and anytime algorithms & Value of computation; stopping and quality--time trade-offs & Uses these principles to assess which allocation conclusions the reviewed experiments support.\\
Training--inference allocation & Model size, training duration, serving demand, and inference effort & Asks that a training option and an inference option be scored on the answer actually returned and costed over the whole system, verifier and index included.\\
Retrieval and datastore scaling & Effects of external memory on loss and task accuracy & Separates new information, re-access to seen information, and processing cost as distinct explanations of a gain.\\
Compound-system optimization & Selection and optimization of models, prompts, modules, and calls & Asks which configuration gains survive when both configurations are costed the same way and allowed the same information.\\
Reasoning and test-time surveys & Taxonomies of inference, learning, verification, efficiency, and agent architectures & States what each experimental design can and cannot show about where to spend.\\
Resource-aware agent evaluation & Cost--quality trade-offs, holdouts, and reproducibility & Records in one place what the agent was allowed, what it consumed, what it could see, and when it stopped.\\
\bottomrule
\end{tabularx}

\end{table}

\section{A framework for interpreting system scaling}
\label{sec:framework}
The three mismatches of the introduction are questions about a score. Answering them needs a description of the system that produced the score. The example has three parts to describe. One is the budget and how it was split between building and running. One is the procedure that spent it, including the rule that picked the returned answer. One is what the system was allowed to know. The framework below keeps these three parts apart and calls them budgets, mechanisms, and information. The three parts are not the three mismatches. The mismatches are questions asked about a score from such a system: how it was counted, what the system knew, and which costs were charged. The framework then distinguishes the performance of a tested procedure from the best performance attainable under a budget and connects development expenditure to the cost of serving queries. Finally, it explains why the choice of score can change an apparent interaction between resources. These distinctions guide the comparisons in Section~\ref{sec:synthesis}.

\subsection{Budgets, mechanisms, and information}
The framework describes a system by three things: what it spends, how it spends it, and what it is allowed to know. Quality is then written as a function of the three. Let $k$ specify a task distribution, scoring rule, and evaluation protocol. Let $\mathbf b$ collect resource budgets, $\mathbf s$ specify mechanisms for using them, and $\I$ describe the information made available. Expected quality, oriented so that larger is better, is
\begin{equation}
Q_k=g_k(\mathbf b,\mathbf s;\I),\qquad (\mathbf b,\mathbf s)\in\F_k(\I).
\label{eq:surface}
\end{equation}
The review calls $g_k$ the capability surface and $\F_k(\I)$ the feasible set, the budget--mechanism pairs that can actually be run under the available information. The capability surface is a name for how quality responds to resources in one setting. Budgets can include training computation, inference computation, storage, tool calls, and environmental steps. Mechanisms include the architecture, data-acquisition policy, retrieval rule, verifier, coordination structure, and stopping policy. Verifier is used here for any component that judges candidates, which some studies call proposals. Where the difference matters, a selector picks one candidate from several and a checker accepts or rejects candidates one at a time. Information access specifies the corpus, tool outputs, permissible observations, feedback, and prior assumptions. It has a qualitative component: a million more observations and one experiment that settles which of two causal stories is true are not interchangeable quantities.

These distinctions prevent several misleading comparisons. Increasing an existing datastore changes a budget and its content; replacing its embedding model changes a mechanism. Allowing the system to consult an answer key, an oracle, changes information access even if the cost of consulting it is small. Increasing an agent's token allowance differs from training a policy that can use that allowance. A retrieval system can expose new information or make previously encountered information more accessible, and these effects require different controls.

The feasible set $\F_k(\I)$ also records dependencies among budgets. For a specified dense training recipe, training compute is approximately proportional to parameter count times processed tokens \citep{hoffmann2022training}; the three are not independent experimental axes. Model depth and width, token repetition, and data quality also matter. A hardware memory limit can rule out a large model regardless of its estimated FLOPs, while an interactive task can impose a deadline that prevents serial search.

A pretraining scaling study measures a restricted part of Eq.~\eqref{eq:surface}. Three conditions are especially useful to record. The first is the acquisition policy supplying additional training data. The second is the inference policy converting inputs into outputs. The third is the configuration of memory, tools, verification, and coordination. A policy can be adaptive while remaining fixed across a scaling comparison. Holding these conditions fixed makes a result interpretable, but leaves the return to changing them unmeasured.

\subsection{Performance curves, frontiers, and what a budget buys}
A performance curve measured under one procedure is different from the best that any feasible procedure could attain. Under an upper-bound budget $\mathbf B$, define the attainable frontier
\begin{equation}
Q_k^*(\mathbf B;\I)=\sup\{g_k(\mathbf b,\mathbf s;\I):
(\mathbf b,\mathbf s)\in\F_k(\I),\ \mathbf b\leq\mathbf B\}.
\label{eq:frontier}
\end{equation}
The supremum need not be attained by any single feasible configuration; the frontier is the least upper bound on what feasible configurations can achieve. If extra resources are optional and earlier systems remain feasible, this frontier is nondecreasing as an allowance expands. An observed procedure can nevertheless deteriorate with more agents, longer context, or additional search. Such a decline identifies a property of that procedure and setting; it does not show that every feasible use of the additional resource is harmful. Conversely, the expected quality of any tested feasible configuration cannot exceed the attainable frontier, and its empirical estimate is subject to sampling uncertainty. The best tested configuration should not be described as globally optimal without an appropriate proof or search guarantee.

The relevant objective often spans building and running. From here on, building is called development and running is called run-time. For a deployment horizon of $N$ queries, an additive accounting model is
\begin{equation}
C_{\mathrm{life}}=C_{\mathrm{dev}}+C_{\mathrm{update}}(N)
 +\sum_{t=1}^{N}\E[C_{\mathrm{run},t}],
\label{eq:lifecycle}
\end{equation}
where development can include training, verifier construction, indexing, and configuration search. Updates include subsequent model or index maintenance. The identity holds once the boundary is stated: which costs are charged and which are left out. The review calls that list the accounting boundary. The identity is not a claim that workloads are stationary. A stationary simplification replaces the sum by $N\E[C_{\mathrm{run}}]$. Shared pretraining expenditure can be treated as sunk for one deployment decision but charged for an end-to-end technology comparison; both views are useful if reported separately.

FLOPs, tokens, monetary charges, energy, and latency are not interchangeable. Converting them requires assumptions about hardware, batching, utilization, and prices. Multiple constraints often remain: minimize lifecycle expenditure subject to a quality target, a latency quantile, a memory limit, and an error-risk requirement. A single cost--quality plot is informative only to the extent that these other constraints and its accounting boundary are explicit.

Efficiency methods also change what a budget buys. Distillation transfers behavior from a teacher to a smaller student, moving expenditure toward development in exchange for a potentially cheaper deployed model \citep{hinton2015distilling,gu2024minillm}. It does not guarantee preservation of every capability. Quantization changes numerical precision and memory requirements: LLM.int8(), GPTQ, and AWQ illustrate different routes to reducing inference resource demands while controlling degradation in their evaluated settings \citep{dettmers2022int8,frantar2023gptq,lin2024awq}. These interventions can change both attainable quality and the feasible hardware configurations. Comparisons should therefore specify precision, calibration data, kernels, and the cost of producing the compressed model, rather than treating a parameter count as a fixed serving expense.

Exact speculative decoding offers a distinct case. A draft model proposes tokens that the target model checks with a distribution-preserving acceptance procedure \citep{leviathan2023speculative,chen2023speculative}. Under that procedure's assumptions, acceleration need not change the target output distribution; its benefit depends on draft acceptance and execution costs, not an increase in reasoning ability. Likewise, PagedAttention changes serving feasibility through key--value cache management and sharing \citep{kwon2023pagedattention}. A latency or memory improvement can permit more candidates within an allowance even when quality per candidate is unchanged. Consequently, scaling comparisons should report both the reasoning policy and the implementation that converts a budget into usable computation.

\subsection{Interactions depend on the outcome being measured}
One allocation question is whether increasing one resource makes another more useful. For example, does retrieval improve a larger model more than a smaller one? The answer depends on how improvement is measured. Holding mechanisms and information access fixed, a smooth response function expresses this interaction through the mixed partial $g_{ij}$, where $g_i=\partial g/\partial b_i$ and $g_{ij}=\partial^2g/\partial b_i\partial b_j$. Even an increasing transformation of the score can change the sign of this interaction. For $\widetilde g=h(g)$,
\begin{equation}
\widetilde g_{ij}=h'(g)g_{ij}+h''(g)g_i g_j.
\label{eq:metric}
\end{equation}
For example, $g=b_1b_2$ has positive mixed partial, whereas $h(g)=-1/g$ is increasing on $g>0$ and gives a negative mixed partial. Both rank every allocation identically. Thus ``complementarity'' inferred from a cross-partial must name its score scale.

By contrast, an increasing transformation preserves level sets and their local substitution rate, $db_j/db_i=-g_i/g_j$ where defined. Cost to achieve a fixed performance level can therefore provide a more decision-relevant comparison. This invariance does not apply to switching between objectives that are not monotone transformations of one another. For a question with one correct answer, one metric can score the probability the model assigns to that answer, and another can score only whether the model chooses it. Perplexity and classification accuracy are such a pair, and they can rank systems differently. The broader lesson is consistent with debates about apparent emergence under different metrics \citep{wei2022emergent,schaeffer2023emergent}: a numerical shape needs an explicit measurement interpretation.

Mechanism changes are assessed by contrasting the capability surface under one mechanism with the surface under another, not by differentiating with respect to a label. A retriever or verifier can change which surface is attainable, while a new observation can change $\I$. Neither operation should be conflated with moving along a fixed curve.

\section{Critical synthesis of the scaling evidence}
\label{sec:synthesis}
The framework directs attention to four questions. Each is a different kind of claim, and the evidence tables below are read against them.
\begin{enumerate}[label=(\roman*),leftmargin=*]
\item \emph{Conditional performance:} Does a specified procedure improve when a resource changes? The claim concerns that procedure under the stated evaluation conditions.
\item \emph{Mechanism:} Which component or resource accounts for the change? Attribution requires a design that separates it from other changes.
\item \emph{Tested allocation:} Which evaluated system is preferable under a common objective and operating requirements? The comparison must align outcomes, access permissions, accounting boundaries, and constraints.
\item \emph{Frontier:} What can be established about the best feasible performance, including alternatives that were not tested? This requires a guarantee extending beyond the evaluated configurations.
\end{enumerate}
The review asks these questions of studies on pretraining, inference, retrieval, verification, and agent systems. Each table records what was varied, what was measured, and how far the result supports an allocation conclusion. Section~\ref{sec:jointsupport} brings the answers together.

\subsection{Pretraining and the split between development and run-time}
\label{sec:pretraining}
Empirical scaling studies established useful regularities well before current reasoning models \citep{hestness2017scaling}. The influential language-model analyses of \citet{kaplan2020scaling} and \citet{hoffmann2022training} relate predictive loss to model and data scale, but differences in experimental ranges and fitting procedures can change the allocation they recommend. Scaling theory also distinguishes regimes and possible mechanisms behind power-law behavior \citep{bahri2024explaining}. A fitted relationship should consequently be interpreted within its training recipe, data regime, and extrapolation range rather than treated as an architecture-independent property of intelligence.

Data quantity is particularly easy to misinterpret. Unique information, processed tokens, and repeated exposure are different resources. \citet{muennighoff2023data} study data-constrained training and the returns to repetition, while \citet{villalobos2024data} analyze constraints on the supply of human-generated text. Neither licenses treating all additional tokens as equally informative. The choice between acquiring new data, reweighting existing data, and repeating it changes the acquisition policy as well as its expense. Evaluation loss on the pretraining distribution must also be separated from downstream transfer, whose gains require their own measurements.

Serving demand changes the optimization problem even when predictive loss remains the target. \citet{sardana2024beyond} incorporate inference into scaling calculations, examine monetary as well as compute costs, and train 47 models to probe long training regimes. Their analysis favors smaller, longer-trained models under substantial inference demand and shows that fits based on conventional token-to-parameter ratios can overestimate returns in more extreme regimes. This is direct prior work on lifecycle allocation, not merely evidence that smaller models can be useful.

Task-specific inference introduces another dimension. \citet{jones2021boardgames} examines training and test-time search in board games. \citet{roberts2026overtraining} jointly model training scale and inference sampling. They fit relationships for loss and for pass@$k$, the probability that at least one of $k$ candidates passes the task's test. Pass@$k$ is coverage under that test. Their analysis supports an allocation argument for the measured coverage objective under its compute approximation. It does not by itself establish the cost of reliably selecting a successful sample in deployment. This distinction is especially consequential when generated lengths vary or selection requires a separate model.

Table~\ref{tab:compute} summarizes the development-side studies above together with the inference-policy studies of Section~\ref{sec:inference}. The evidence supports a shift from training-only optimization toward task- and workload-specific allocation. It does not establish one common exponent or a universally preferable balance between parameter scale and inference effort.

\subsection{Test-time computation is a policy, not just a token count}
\label{sec:inference}
Inference scaling includes several mechanisms with different cost structures. Chain-of-thought prompting exposes intermediate steps \citep{wei2022chain}. Self-consistency aggregates sampled answers \citep{wang2023selfconsistency}. Tree of Thoughts searches over intermediate states \citep{yao2023tree}. These procedures can use comparable numbers of tokens while differing in serial depth, branching, selection, and feedback. Parallel sampling can reduce elapsed time when hardware permits, whereas a serial dependency cannot generally be parallelized in the same way. Reporting only total tokens hides that distinction.

\citet{snell2024scaling} compare verifier-guided search and learned revisions and show that the preferable tested strategy depends on prompt difficulty. Their training--inference comparison is conditional on the smaller model having useful baseline success and on the accounting boundary. The study also considers estimating difficulty rather than assuming it is known. The transferable lesson is the importance of matching policy to difficulty, not that a fixed amount of inference can always replace a larger model.

\citet{wu2025inference} directly compare model sizes and inference strategies, including voting, best-of-$N$, and tree search, under compute budgets on mathematical reasoning tasks. Their REBASE tree search spreads expansion across branches in proportion to their reward scores. It illustrates how changing the inference algorithm can shift the observed cost--accuracy trade-off, with smaller models outperforming larger alternatives in tested configurations. This is evidence for joint model--policy selection within those experiments. The comparison remains conditional on the task distribution, available models, scoring mechanisms, and FLOP accounting; it is not a conversion factor between parameters and inference computation.

Call count need not even yield a monotone response for a fixed compound procedure. \citet{chen2024calls} analyze voting and filter-then-vote systems and show how additional calls can improve queries whose answer distribution favors the correct answer while reinforcing errors on other queries. Mixing such queries can produce non-monotone aggregate accuracy. The decline belongs to the tested procedure, not to the allowance. It is a concrete instance of the distinction in Eq.~\eqref{eq:frontier}.

Training changes which inference policies are productive. DeepSeek-R1 demonstrates reasoning-oriented reinforcement learning \citep{deepseek2025r1}; s1 combines a curated reasoning dataset with budget forcing, which makes the model stop or keep thinking at a set token count \citep{muennighoff2025s1}. These are changes in the production of a reasoning policy, not evidence that extending any model's output will have the same effect. Earlier adaptive-computation architectures likewise make computation allocation part of the mechanism \citep{graves2016act,dehghani2019universal}. A budget-only ablation should therefore be distinguished from an intervention that retrains or prompts the system to use a budget differently.

Metareasoning makes this distinction actionable: continue only when the expected gain warrants the incremental expense. \citet{desabbata2024rational} train selective reasoning with a cost-sensitive objective, while \citet{kalayci2025stopping} connect inference-time selection to optimal stopping. These approaches formalize part of the allocation problem already. Their practical application still depends on estimating task difficulty, the usefulness of another sample, or the calibration of a reward signal. The uncertainty in these estimates should accompany any claimed savings.

\begin{table}[!htbp]
\centering\small
\caption{Selected evidence on training and inference allocation. Statements summarize the reported designs; the final column states the limits of the allocation inference.}
\label{tab:compute}
\begin{tabular}{@{}L{2.8cm}L{3.1cm}L{3.9cm}L{4.7cm}@{}}
\toprule
Study & Main variation & Outcome and supported inference & Boundary or qualification\\
\midrule
\citet{hoffmann2022training} & Parameters and training tokens & Loss-based compute allocation within the studied training regime & Does not optimize retrieval, selectors, or deployment demand.\\[5pt]
\citet{sardana2024beyond} & Training duration, size, and serving demand & Inference-aware development changes the cost-optimal model choice & Depends on loss fits, serving assumptions, and the cost model.\\[5pt]
\citet{snell2024scaling} & Search, revisions, and prompt difficulty & Allocation among tested inference strategies can improve compute efficiency & Difficulty estimation, trained proposer/verifier, and task regime matter.\\[5pt]
\citet{wu2025inference} & Model size and inference algorithm & Smaller models with suitable sampling or search can improve tested cost--accuracy trade-offs & Mathematical reasoning results and FLOP accounting do not establish universal model substitution.\\[5pt]
\citet{chen2024calls} & Call count under voting and filtering & More calls can help some queries and hurt others, producing non-monotone aggregate returns & A procedure's performance curve is distinct from the attainable frontier over an expanding feasible set.\\[5pt]
\citet{brown2024monkeys} & Number of generated candidates & Coverage continues improving in settings where practical selection is weaker & Pass@$k$ requires a success criterion; coverage is not returned-answer accuracy.\\[5pt]
\citet{muennighoff2025s1} & s1: curated reasoning training and budget forcing & A trained model can use a controlled reasoning budget & Policy and training changes accompany token-budget comparisons.\\[5pt]
\citet{desabbata2024rational} & Cost-sensitive reasoning policy & Learned selective reasoning improves the measured quality--token trade-off & Training the allocation policy has a separate development cost.\\[5pt]
\citet{roberts2026overtraining} & Parameters, training tokens, and sample count & Joint allocation can favor overtraining for fitted pass@$k$ objectives & Simplified inference cost and coverage objective do not price a full selector.\\[5pt]
\citet{kalayci2025stopping} & Adaptive stopping versus fixed sample counts & Stopping policies can improve reward--generation trade-offs & Reward model, distributional assumptions, and calibration define the guarantee.\\
\bottomrule
\end{tabular}
\end{table}
\FloatBarrier

\subsection{Retrieval changes both accessibility and resource accounting}
\label{sec:retrieval}
Retrieval systems distribute knowledge between parameters and external stores. Nearest-neighbor language models, REALM, RAG, RETRO, and Atlas provide different mechanisms for accessing external evidence \citep{khandelwal2020knnlm,guu2020realm,lewis2020rag,borgeaud2022retro,izacard2023atlas}. Their architectures and training requirements differ, so datastore tokens cannot be treated as an interchangeable substitute for training tokens across methods. Retrieval quality depends on the index, query representation, corpus composition, and the ability of the model to use the retrieved material.

\citet{shao2024datastore} systematically scale a datastore to 1.4 trillion tokens and report gains in language modeling and knowledge-intensive tasks. They also compare model, training-data, and datastore choices under a training-compute perspective. These results establish datastore size as an important resource axis; converting that advantage into a lifecycle claim additionally requires indexing, retrieval, context-processing, storage, and workload assumptions. The qualification concerns the interpretation of the accounting boundary, not a claim that the study disregards engineering efficiency.

Two recent studies sharpen the role of reuse. \citet{fang2025reuse} retrieve from pretraining data and observe gains that persist after decontamination, showing that parametric training can leave useful information accessible through another mechanism. Their reported gain, expressed as the extra pretraining compute that would buy the same performance, is relative to a pretraining-only performance curve, rather than an end-to-end monetary saving. \citet{singh2026memorize} vary model size, pretraining, and retrieval data, distinguishing reused from held-out corpora. They report different model-scale patterns for the perplexity of the correct answer and decision accuracy. Their appendix explicitly estimates embedding, retrieval-related inference, and storage costs; token-allocation findings should not be mistaken for a fully optimized lifecycle comparison.

Together, these studies support three different explanations for a retrieval gain: exposure to previously unavailable information, more effective access to information already encountered, and extra computation for processing evidence. A convincing experiment separates them using corpus-overlap controls, retriever and context ablations, and one accounting boundary. Larger models can need less factual assistance yet use retrieved evidence more effectively on a decision metric. These possibilities can coexist, so a single label such as ``retrieval substitutes for scale'' is insufficient. Table~\ref{tab:retrieval} summarizes the relevant distinctions.

\begin{table}[!htbp]
\centering\small
\caption{Retrieval evidence is conditional on both the information exposed and the mechanism that makes it usable. Mechanism-family references appear in the text above.}
\label{tab:retrieval}
\begin{tabularx}{\textwidth}{@{}L{3.1cm}YY@{}}
\toprule
Study or family & What the evidence establishes & What is needed for an allocation claim\\
\midrule
RAG, RETRO, Atlas & External evidence can be integrated through distinct retrieval and training mechanisms & Match corpus access, reader-model training, and context costs before comparing scale.\\[4pt]
\citet{shao2024datastore} & Larger datastores improve measured outcomes in the studied retrieval systems & Extend training-compute comparisons with index, serving, storage, and update costs.\\[4pt]
\citet{fang2025reuse} & Reaccessing pretraining data can improve task accuracy after decontamination & Keep pretraining-equivalent gains separate from full-system cost savings.\\[4pt]
\citet{singh2026memorize} & Reuse and held-out retrieval differ; scale interactions vary by outcome & Interpret perplexity and accuracy separately and incorporate the reported cost components.\\
\bottomrule
\end{tabularx}

\end{table}
\FloatBarrier

\subsection{Generation, verification, and the returned answer}
Repeated generation increases opportunities for success, but evaluation must specify how a system recognizes that success. Code-generation evaluation popularized pass@$k$ as the coverage measure \citep{chen2021codex}. Training verifiers and process reward models makes selection itself a learned component \citep{cobbe2021verifiers,lightman2023verify}. \citet{brown2024monkeys} demonstrate the importance of repeated sampling while also exposing limitations of practical selection. When at least one candidate is correct, the selector must still pick it. The probability that it does so is the conditional selection accuracy. Together, these results motivate separate measurements of coverage, conditional selection accuracy, and returned-answer accuracy.

Verification mechanisms provide different guarantees. A proof checker can certify a formally specified statement, while a learned reward model predicts quality with possible distribution-dependent errors. Unit tests verify the tested behaviors rather than every possible behavior of a program. A benchmark answer key supplies privileged information if a deployed system would not receive it. The neural and symbolic components of AlphaGeometry illustrate how generation and explicit deduction can be combined within a specified problem domain \citep{trinh2024alphageometry}. The LLM-modulo proposal similarly pairs a generating language model with an external critic \citep{kambhampati2024llmmodulo}.

Selection also changes the distribution presented to a verifier. \citet{huang2025bestofn} formalize how imperfect reward models and the base policy's coverage of high-quality outputs govern inference-time alignment. Best-of-$N$ can be optimal under stringent coverage conditions with an appropriately chosen $N$. Increasing $N$ further can select outputs that score well on the reward model but are worse in fact. This is called reward hacking, and it can reduce true reward. Their pessimistic selection procedure addresses this overoptimization under specified assumptions. Thus, a verifier's average error on independently sampled outputs need not characterize its behavior inside optimization. The relevant allocation quantity is the joint performance of generator, search, and verifier on the candidates they actually produce; verifier training and evaluation are surveyed by \citet{venktesh2025verification} and \citet{liu2026reward}.

The training signal also affects what inference scaling can achieve. \citet{setlur2025verification} compare verifier-free learning from expert traces with verifier-based approaches that use rewards on sampled trajectories. Their theoretical separation depends on how varied the sampled trajectories are, on how rewards are spread among them, and on how training data grows with the inference horizon. Experiments examine related behavior in the studied models. This strengthens the case for treating verification as a resource and mechanism, while qualifying a simple generation-versus-checking interpretation: the comparison changes supervision and learning, not only the number of checks applied to one fixed generator. Verifier development and training must therefore be included when translating the result into an allocation recommendation.

Self-correction is especially sensitive to the source of feedback. \citet{huang2024selfcorrect} find limitations of intrinsic self-correction in the reasoning settings they study; this does not establish an impossibility theorem for all models or tasks. Feedback from an executable environment, a human, an answer key, and the model's own critique must be distinguished. Increasing critique tokens without introducing useful information or a better checking mechanism can leave the active bottleneck unchanged.

\subsection{Interaction, coordination, and evaluation infrastructure}
Agentic systems expose the importance of mechanisms because they combine multiple calls, observations, and state transitions. ReAct interleaves reasoning and action, Toolformer learns tool-use opportunities, and Reflexion incorporates feedback through verbal memory \citep{yao2023react,schick2023toolformer,shinn2023reflexion}. These methods establish useful constructions; their task gains do not constitute a general law for scaling tool calls or memory. Each additional call has both an information role and a cost, and may introduce a new failure mode.

\citet{shen2025interaction} compare longer interaction with more per-step computation and develop training that supports longer horizons. Their web-agent evidence makes clear why acting can provide information unavailable to further internal deliberation. It also links productive run-time scaling to development: a longer interaction allowance and a policy trained to exploit it are distinct interventions. \citet{kim2025scalingagents} examine different coordination architectures across a large set of configurations and show task-dependent effects. Detailed cost tracking covers only a subset of their configurations, so cost conclusions rest on fewer runs than quality conclusions. Agent count alone consequently fails to determine either cost or quality.

The evaluation literature is already moving toward system comparisons. \citet{kapoor2024agents} emphasize joint optimization and cost--accuracy trade-offs, and the Holistic Agent Leaderboard develops infrastructure for broader agent evaluation \citep{kapoor2025hal}. \citet{mcfadyen2026inference} examine how inference allowances and feedback affect frontier-model evaluation. A task counts as solved if it succeeds at any point up to the allowance, and the share of tasks solved this way is cumulative success. They report curves of cumulative success and of score against the allowance. These curves should be read under the declared oracle-feedback and no-feedback protocols, rather than automatically interpreted as the accuracy of a final answer selected without a model acting as judge. Their reported token accounting excludes the tokens of the judging model, an explicitly stated accounting boundary relevant to end-to-end costing. Table~\ref{tab:agents} records, for each verification and agent study, the limit on its allocation conclusion.

\begin{table}[!htbp]
\centering\small
\caption{Selected verification, system, and evaluation studies. The last column distinguishes the observed result from a broader claim about reliable deployment.}
\label{tab:agents}
\begin{tabularx}{\textwidth}{@{}L{3.2cm}YY@{}}
\toprule
Study & Relevant system intervention & Limit on the allocation conclusion\\
\midrule
\citet{lightman2023verify} & Learned process verification & Supervision and verifier development are resources, not free annotations.\\[4pt]
\citet{setlur2025verification} & Verifier-based versus verifier-free learning and inference & The separation depends on policy and reward assumptions, supervision, and joint training--inference scaling.\\[4pt]
\citet{huang2025bestofn} & Best-of-$N$ and pessimistic reward-guided selection & Coverage and reward-model error govern scaling; increasing $N$ can amplify reward hacking.\\[4pt]
\citet{chen2025selector} & Model choice across compound-system modules & The optimized object is a bounded module configuration under its scoring assumptions.\\[4pt]
\citet{shen2025interaction} & Longer interaction and curriculum-based training & Added observations and changed training must be distinguished from internal thinking.\\[4pt]
\citet{kim2025scalingagents} & Agent architecture, model, and task configuration & More agents can help or hurt under a particular coordination rule; cost coverage varies by experiment.\\[4pt]
\citet{kapoor2024agents} & Joint agent optimization and cost-aware comparison & Equal comparison requires suitable holdouts, accounting, and tuning opportunities.\\[4pt]
\citet{mcfadyen2026inference} & Token allowance and feedback protocol & Cumulative success, oracle access, and excluded judge costs qualify deployment conclusions.\\
\bottomrule
\end{tabularx}

\end{table}
\FloatBarrier

\subsection{What the evidence jointly supports}
\label{sec:jointsupport}
One finding runs through every literature above: a resource helps only through a mechanism that uses it. More inference needs a policy, more retrieval needs a retriever and a model that can use what it retrieves, and more agents need a rule for passing information among them. Beyond that, the three mismatches of the introduction recur in each literature. First, how success was counted. Coverage, returned-answer accuracy, loss, and cumulative success answer different practical questions, and the studies above report different ones. Second, what the system was allowed to know. Answer keys used to sort questions by difficulty and oracle feedback during a run both arise in the reviewed evidence. The evaluation guidance above asks for held-out sets for the same reason. Each of these changes what a score means for a deployed system. Third, what was counted as cost. A verifier, an index, a judge, and a configuration search each cost something, and several studies leave one of them outside the boundary.

Together, these conclusions sharpen the four questions that opened this section. A study can answer one of them without answering the others, and a scaling result identifies whichever of them it establishes. The final columns of the evidence tables state the limit on each result's allocation conclusion; Section~\ref{sec:envelope} translates the distinctions into reporting and comparison requirements.

The next step depends on the claim being considered. A component comparison may need a controlled intervention; a deployment comparison may need additional cost measurement or held-out system selection. A claim about untested feasible alternatives also needs a justification, such as an analytical bound or an appropriate search guarantee. Naming the missing step keeps the original result and says what to run next.

\section{Capability bottlenecks: computation, information, and state}
\label{sec:bottlenecks}
Knowing that one procedure outperforms another does not yet say which resource to change next. That decision depends on what is limiting performance on the task. The three kinds of task named in the introduction each expose a different limit. A task that produces a result that must then be checked is called discovery here, and its limit is validation. A task that predicts what an intervention would do is causal inference, and its limit is evidence that distinguishes competing explanations. A task that carries out a long chain of steps is sequential composition, and its limit is keeping intermediate results available for later computation. Each case motivates a different experimental intervention. They illustrate how task analysis guides resource allocation.

\subsection{Discovery requires a claim about validation}
Discovery tasks ask a system to produce a result not explicitly supplied to it and to establish an appropriate basis for accepting that result. This is an operational description. It does not by itself rule out training contamination; novelty needs a separate provenance argument, and importance needs more than novelty. For allocation, the central distinction is between finding a candidate and validating it.

Three bottlenecks can arise. A weak generator may rarely propose a correct candidate; a weak selector may fail to recognize a candidate already present; or a costly checker may make broad search uneconomical. More sampling principally addresses the first. Better supervision or checking can address the second. Changes to experimental design, batching, or surrogate screening may reduce the third, but can introduce additional uncertainty. Formal verification can offer a strong acceptance criterion for a specified formal object, whereas empirical discovery may require new measurements. The two should not share an unqualified ``verified'' label.

The allocation consequence is to measure the complete acceptance procedure. A candidate-generation curve paired with an oracle the deployed system will not have estimates an opportunity for further work, not an achieved deployable system. Conversely, a valid checker can make a modest generator useful when repeated proposals are inexpensive. Whether generation and verification are substitutes or complements depends on the acceptance target, their errors, and their respective development and run-time costs.

\subsection{Causal identification is an information constraint}
Causal inference asks what would happen under a specified intervention. Answering needs a causal model, or assumptions, under which the data single out one answer. Causal inference calls this identification. Observational fit alone need not determine that answer. This distinction is foundational in causal inference \citep{verma1990equivalence,pearl2009causality,peters2017elements,bareinboim2022hierarchy}. It therefore serves as a diagnostic counterexample to the claim that additional internal computation can always compensate for limited information.

Here $\operatorname{do}(X=x)$ denotes an intervention that sets $X$ to $x$. Consider two structural causal models with independent standard normal disturbances $U,V$ and $|\rho|<1$:
\begin{align}
\mathcal M_A:&\quad X=U,\qquad Y=\rho X+\sqrt{1-\rho^2}\,V,\\
\mathcal M_B:&\quad Y=V,\qquad X=\rho Y+\sqrt{1-\rho^2}\,U.
\end{align}
Both generate the same observational bivariate normal distribution, with correlation $\rho$. Yet they imply $\E_A[Y\mid\operatorname{do}(X=x)]=\rho x$ and $\E_B[Y\mid\operatorname{do}(X=x)]=0$. An estimator given only observational samples, with the same auxiliary information and independent internal randomization in both worlds, has the same output distribution in each. For estimating this interventional mean, its worst-case mean squared error across the two worlds is at least $(\rho x)^2/4$; Appendix~\ref{app:derivations} gives the elementary argument.

A requirement of worst-case mean squared error strictly below $(\rho x)^2/4$ is infeasible under this information restriction, regardless of the observational sample size or internal computation budget. An allocation comparison for that requirement must therefore consider information or assumptions that distinguish the two worlds. This is an application of nonidentification to feasibility, not a new causal theorem or a blanket limitation on language models. A language model may already encode descriptions of experiments, temporal constraints, or domain assumptions that distinguish the two worlds. If so, it does not operate under the example's information restriction. The relevant experiment must specify what evidence is available and then vary observations, identifying assumptions, or interventions separately. Work on the number and design of interventions provides established foundations for this acquisition problem \citep{eberhardt2005experiments}.

The distinction also clarifies when retrieval and interaction matter. Retrieving more samples from the same observational distribution may reduce estimation error without resolving ambiguity. Retrieving a valid experimental report can change identification. Interacting with an environment can expose useful causal information, but only if the action, observation, and assumptions support the target query. An interaction count alone does not establish that contribution.

\subsection{Sequential composition depends on accessible computation and state}
Tasks such as executing a program, constructing a proof, or navigating an environment require intermediate results to remain available to later operations. Relevant resources include architectural depth, generated steps, recurrent state, context, and external memory. Theoretical analyses of transformers formalize distinctions between parallel and serial computation under specified precision and depth assumptions \citep{merrill2023parallelism,merrill2024cot,li2024serial,feng2023cot}. More recent depth results further qualify what follows from a constant-depth restriction \citep{merrill2025depth}.

These results hold within their computational models. A lower bound for one fixed-depth family does not show that larger or deeper models cannot solve a task. Expressibility does not guarantee that training will find the required algorithm. Empirical compositionality studies expose failures on particular task families \citep{dziri2023faith}, while algorithmic-generalization evaluation asks how task complexity changes demands on a system \citep{ito2025quantifying}.

An informative scaling experiment varies task depth or length alongside the resources that can supply sequential computation and retain state. It should separate the number of generated tokens from their causal contribution: a longer trace can contain repetition or ineffective deliberation. External memory may help when context is limiting, but it is not universally necessary if equivalent state remains accessible elsewhere. Likewise, adaptive inference can be efficient without being required for every task that admits a fixed computation schedule.

Table~\ref{tab:bottlenecks} connects these task demands to experimental interventions. Its categories overlap: a discovery task can require causal identification and long sequential reasoning. The first row adds the pretraining case of Section~\ref{sec:pretraining} for comparison, under the label predictive competence. Their purpose is to identify the binding constraint, rather than to partition tasks.

\begin{table}[!htbp]
\centering\small
\caption{Task demands help determine which resource intervention to test. A listed resource is a candidate remedy under stated conditions, not a universal prerequisite.}
\label{tab:bottlenecks}
\begin{tabularx}{\textwidth}{@{}L{2.6cm}YYY@{}}
\toprule
Task demand & Candidate bottleneck & Discriminating intervention & Primary measurement\\
\midrule
Predictive competence & Insufficient training or distributional coverage & Vary model/data under a fixed recipe; separately vary acquisition & Held-out loss and task transfer, each named\\[4pt]
Discovery & Poor proposals, weak selection, or expensive checking & Cross generator and verifier budgets; vary access to valid feedback & Coverage, returned-answer accuracy, and validation cost\\[4pt]
Causal inference & Missing identifying information & Compare observations, justified assumptions, and interventions & Error on identified targets; sensitivity to assumptions\\[4pt]
Sequential composition & Insufficient sequential computation or accessible state & Cross task depth with model depth, reasoning steps, and memory & Success versus task complexity and consumed resources\\
\bottomrule
\end{tabularx}

\end{table}

\section{Worked analytical examples}
\label{sec:examples}
The task cases above say which resource to test once the limiting constraint is known. The examples below take the next step and compare specific allocations under a stated task requirement. The first compares a retrieval system with a larger model. The second compares more candidates with a better selector. The third compares stopping rules. All probabilities and costs are stipulated for illustration. They are not fitted to the reviewed studies, do not reproduce their experiments, and do not estimate a deployed system's performance.

\subsection{Retrieval: metric choice and deployment volume}
\label{sec:retrievalexample}
Suppose a one-token, two-answer task has correct answer A. A small model assigns A probability $0.35$ without retrieval and $0.45$ with retrieval. A large model assigns probabilities $0.49$ and $0.55$, respectively. Table~\ref{tab:metricexample} reports the consequences. On this single task, gold-answer perplexity is $1/P(A)$, while deterministic classification chooses the more probable answer.

\begin{table}[!htbp]
\centering\small
\caption{Hypothetical probabilities, not experimental measurements. Retrieval reduces perplexity more for the small model but improves accuracy only for the large model.}
\label{tab:metricexample}
\begin{tabular}{@{}lrrrr@{}}
\toprule
Configuration & $P(A)$ & $P(B)$ & Perplexity & Accuracy\\
\midrule
Small, no retrieval & 0.35 & 0.65 & 2.857 & 0\\
Small, retrieval & 0.45 & 0.55 & 2.222 & 0\\
Large, no retrieval & 0.49 & 0.51 & 2.041 & 0\\
Large, retrieval & 0.55 & 0.45 & 1.818 & 1\\
\bottomrule
\end{tabular}

\end{table}

Retrieval reduces perplexity by approximately $0.635$ for the small model and $0.223$ for the large one. Nevertheless, accuracy improves only for the large model, which crosses the decision threshold. This is not just the increasing-transformation issue in Eq.~\eqref{eq:metric}: accuracy and perplexity are different objectives. The example shows that opposite interaction patterns on the two metrics can arise from the same four probabilities. It does not claim to explain all mechanisms behind the empirical patterns in Section~\ref{sec:retrieval}.

\begin{figure}[!htbp]
\centering
\includegraphics[width=\textwidth]{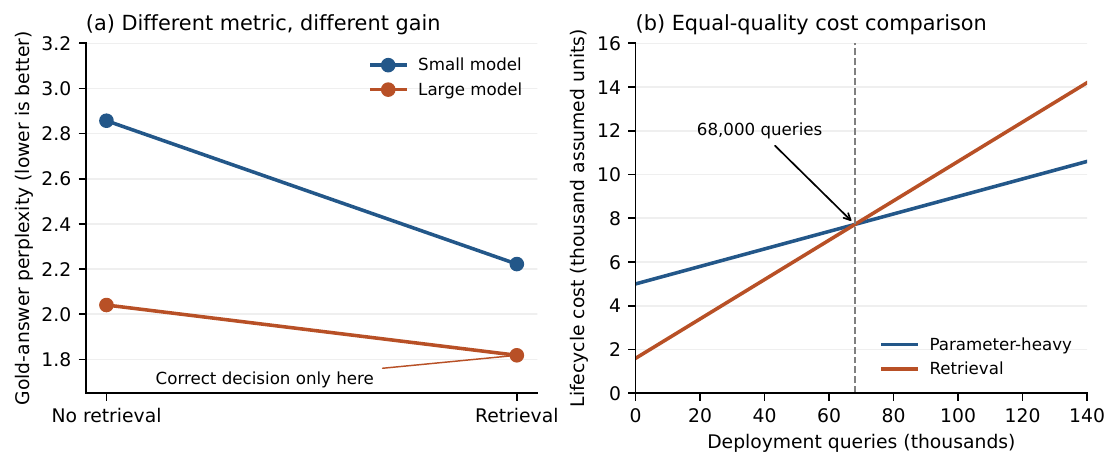}
\caption{\textbf{Two independent analytical illustrations for retrieval comparisons.} (a) The stipulated probabilities in Table~\ref{tab:metricexample} produce a larger perplexity improvement for the small model; only the large model crosses the decision threshold. (b) Under a separate equal-quality assumption, different development and per-query costs reverse the cheaper choice at 68,000 queries. All values are illustrative; panel (b) is not a cost model fitted to panel (a).}
\label{fig:retrieval}
\end{figure}

Now consider a separate, equal-quality cost comparison. Assume two systems meet the same task-quality, information-access, and latency requirements. A parameter-heavy system costs $5{,}000$ units to develop and $0.04$ units per query. A retrieval system costs $1{,}600$ units to develop, including its index, and $0.09$ units per query. Ignoring updates for this illustration, their lifecycle costs are equal at
\begin{equation}
N^*=\frac{5{,}000-1{,}600}{0.09-0.04}=68{,}000\ \text{queries}.
\end{equation}
The retrieval system is cheaper below this volume; the parameter-heavy system is cheaper above it. The assumed task quality is unchanged. Figure~\ref{fig:retrieval} presents this crossing alongside the metric example.

This arithmetic is deliberately simple. Real deployment introduces query mixtures, index updates, cache effects, infrastructure constraints, and uncertain demand. If retrieval provides information unavailable to the other system and necessary to meet the quality target, the parameter-heavy configuration may not be feasible and the comparison no longer applies. Similarly, if monetary charges exclude a provider's development expenditure, they answer a buyer's decision rather than a total technology-cost question.

\subsection{Generation and selection under the same allowance}
\label{sec:selectionexample}
For a fixed task, suppose each independent generated candidate is correct with probability $p$. Coverage after $n$ candidates is
\begin{equation}
C_n=1-(1-p)^n.
\end{equation}
If a selector must return one candidate from the generated set, let $r_n$ be its probability of choosing a correct candidate conditional on the set containing one. Its returned-answer accuracy is exactly
\begin{equation}
A_n=C_n r_n.
\label{eq:selection}
\end{equation}
The decomposition does not require independent candidates; independence is needed only for the displayed expression for $C_n$. Across heterogeneous tasks the correct expression is $\E[C_n(x)r_n(x)]$, where $x$ indexes tasks, not generally the product of the two averages. If the selector can generate a new answer, Eq.~\eqref{eq:selection} describes an incomplete system and must be revised.

Take $p=0.2$ and an allowance of 12 cost units. Each candidate costs one unit to generate. A cheap selector costs $0.1$ units per candidate and is assumed to achieve conditional selection accuracy $0.65$; a stronger selector costs $0.8$ and achieves $0.95$. Assume these conditional accuracies remain constant over the candidate counts considered. The cheap system can generate ten candidates at total cost 11. It obtains coverage $0.893$ and returned-answer accuracy $0.580$. The stronger system can generate six candidates at cost $10.8$. Its coverage is lower, $0.738$, but its returned-answer accuracy is higher, $0.701$.

Thus, under the same allowance, maximizing the number of proposals makes the worse decision in this example. Both systems also leave some allowance unused because candidates are indivisible. A target of $0.70$ returned-answer accuracy is unattainable for the cheap selector under the constant-$r_n$ assumption, regardless of sample count, while the stronger selector reaches it with six candidates. This ceiling is a property of the assumed selector, not a general ceiling on sampling. In practice, $r_n$ can improve or deteriorate with set size, and its estimation requires tasks representative of deployment.

Development costs remain separate. If the stronger selector requires substantial labeling or training, a low-volume deployment might still favor a different configuration. If correctness is available from an executable checker, its acceptance criterion and cost replace the assumed learned-selection behavior.

\subsection{Stopping, abstention, and wrong answers}
\label{sec:stoppingexample}
A different policy generates and checks candidates one at a time, stopping at the first acceptance. Suppose trials are independent and identically distributed. A candidate is correct with probability $p$. The checker accepts a correct candidate with probability $s$ and an incorrect candidate with probability $f$. Thus, acceptance on any trial has probability $a=ps+(1-p)f$. The system makes at most $n$ trials. If none is accepted, it returns no answer, which is called abstention. The outcome probabilities are
\begin{align}
P_{\mathrm{correct}}(n)&=\frac{ps}{a}\big[1-(1-a)^n\big],\label{eq:stopping}\\
P_{\mathrm{wrong}}(n)&=\frac{(1-p)f}{a}\big[1-(1-a)^n\big],\label{eq:wrongreturn}\\
P_{\mathrm{abstain}}(n)&=(1-a)^n,\label{eq:abstention}
\end{align}
for $a>0$. The expected number of attempts is $[1-(1-a)^n]/a$, generally below the cap. Conditional reliability among returned answers is $ps/a$, independent of $n$ under these assumptions. Increasing the cap cannot make abstention more likely. It cannot make a wrong return less likely. For $0<a<1$, it strictly increases wrong-return probability whenever $(1-p)f>0$.

\begin{figure}[!htbp]
\centering
\includegraphics[width=\textwidth]{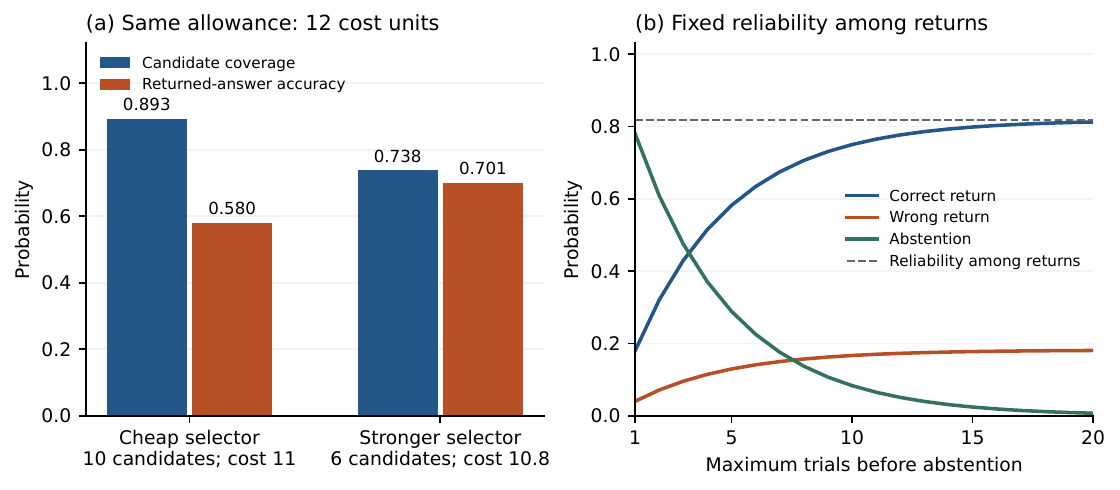}
\caption{\textbf{Analytical generation, selection, and stopping examples.} (a) The assumed stronger selector yields higher returned-answer accuracy with six candidates than the cheap selector with ten, despite lower coverage and similar consumed cost. (b) With the stipulated checker errors, a larger trial cap reduces abstention but increases both correct and wrong returns; reliability conditional on returning stays constant. Lines are calculated from the stated assumptions, not empirical scaling fits.}
\label{fig:selection}
\end{figure}

For $p=0.2$, $s=0.9$, and $f=0.05$, conditional reliability is about $0.818$. Additional trials make a return more likely without improving that reliability. Figure~\ref{fig:selection} shows this behavior alongside the previous allocation example. A benchmark that gives abstentions and wrong answers the same zero score rewards a higher correct-return probability. An application with a separate penalty for wrong answers may prefer another stopping rule. The distinction is a choice of objective, not a paradox.

To decide whether another attempt is worthwhile, the possible outcomes must be valued and the computation charged. Let $v\geq0$ be the value of a correct return. Let $\ell\geq0$ be the loss from a wrong return. Let $c\geq0$ be the cost of generating and checking a candidate. Assign value zero to abstention. If $T_n$ is the number of attempts made under cap $n$, expected utility is
\[
U_n=vP_{\mathrm{correct}}(n)-\ell P_{\mathrm{wrong}}(n)-c\E[T_n].
\]
Increasing the cap by one adds an attempt only if the first $n$ trials were all rejected, an event with probability $(1-a)^n$. The expected net gain from that extra attempt gives
\begin{equation}
U_{n+1}-U_n=(1-a)^n\big[vps-\ell(1-p)f-c\big].
\label{eq:stoppingutility}
\end{equation}
For the stated probabilities, $v=1$, $\ell=5$, and $c=0$, the bracket equals $-0.02$. More trials then increase the probability of a correct return while reducing expected utility, even with no computation cost. Under the homogeneous assumptions, the sign of the gain is the same at every attempt. A stopping policy that responds to how a search develops requires a richer model of changing outcomes, costs, or information.

Adaptive generation, correlated errors, and a checker whose calibration changes during search violate the homogeneous-trial assumptions. The formulas are explanatory baselines, while the reviewed stopping and metareasoning literature addresses richer policy design. Empirical work should estimate the relevant dependence rather than applying an independence formula by default.

\section{A resource envelope for comparable evaluation}
\label{sec:envelope}
The examples in Section~\ref{sec:examples} show why a score alone cannot establish which allocation is preferable. The answer also depends on the information available, the selection and stopping rules, and the resources actually consumed. This section brings those details together in a resource envelope, a structured record of how a result was produced.

\subsection{The score and the process that produced it}
\label{sec:envelopedef}
The envelope has precedents. Holistic Evaluation of Language Models (HELM) emphasizes multiple dimensions and scenarios, and model cards and datasheets document intended use, evaluation conditions, and data provenance \citep{liang2023helm,mitchell2019modelcards,gebru2021datasheets}. Calls to improve AI evaluation reporting similarly stress the interpretation behind aggregate results \citep{burnell2023rethink}. Cost-aware agent evaluation asks for accounting alongside accuracy \citep{kapoor2024agents}. The envelope extends these practices by recording the full procedure that produced a score: development, run-time, information, and selection. Table~\ref{tab:envelope} specifies the fields to report for the comparisons developed here.

Which fields a report must fill depends on what it claims. A lifecycle comparison needs the development and update costs; attribution to a component needs the comparison protocol. If a quantity is unavailable, mark it as unknown. If a component was deliberately left outside the accounting boundary, mark it as excluded. In either case state how the omission affects the comparison, so that the reader can see which conclusions remain open.

\begin{table}[!htbp]
\centering\small
\caption{Operational resource envelope. An unavailable quantity should be marked unknown or excluded, with its effect on the comparison stated; a proxy should identify its conversion assumptions.}
\label{tab:envelope}
\begin{tabular}{@{}L{3.0cm}L{6.2cm}L{5.7cm}@{}}
\toprule
Field & Information to record & Why it changes interpretation\\
\midrule
Task and outcome & Task/version, sampling, split, metric, answer-extraction rule, aggregation, uncertainty & Distinguishes loss, coverage, returned-answer accuracy, cumulative success, and conditional reliability.\\[5pt]
Development & Model/checkpoint; training and post-training; data provenance; index and verifier construction; configuration-search budget & Exposes costs and information incorporated before evaluation, including discarded trials where relevant.\\[5pt]
Run-time allowance & Per-query and global caps for tokens, time, calls, samples, and environment steps & Describes what was permitted, not necessarily what was consumed.\\[5pt]
Run-time consumption & Input/output/reasoning tokens where available; actual calls; latency distribution; memory; hardware or API price/date & Distinguishes actual expense from allowance and makes proxies auditable.\\[5pt]
Information access & Corpus version and overlap; tools; environment observations; resets; labels; human, oracle, or learned feedback & Reveals whether two systems solve the task with comparable evidence and assistance.\\[5pt]
Mechanism & Prompts, retriever, search, branching, coordination, state, verifier, and model-routing rules & Identifies the procedure conditional on which a scaling curve was measured.\\[5pt]
Selection and stopping & Candidate selection; stopping signals; acceptance thresholds; abstention; retries; timeout and failure treatment & Connects generated candidates to the actual answer or action returned.\\[5pt]
Accounting boundary & Charged/excluded components; sunk/shared costs; update schedule; demand horizon; conversion assumptions & Determines whether a saving concerns training, serving, a buyer, or lifecycle resources.\\[5pt]
Comparison protocol & Shared versus separately optimized procedures; tuning opportunity; held-out selection; paired instances and seeds & Separates a controlled intervention from a comparison of the best tested systems.\\
\bottomrule
\end{tabular}
\end{table}

Allowance and consumption need separate reporting. A system with a large cap can stop early; another can consume its entire cap without improving. Means alone can hide rare expensive failures, so tail latency, timeout frequency, and the distribution of consumed resources may affect feasibility. Parallel inference further requires elapsed time and hardware allocation alongside aggregate work. Reporting a zero score for a timeout without its consumed resources understates the expense of failure.

Information access deserves equal prominence. An experiment can legitimately provide an oracle to estimate assisted capability, provided the result is labeled accordingly. It cannot silently transfer that interpretation to autonomous deployment. The same applies to retrieval from test solutions, environmental resets, human corrections, and judge-mediated stopping. Development access also matters: a policy tuned repeatedly on a test set has acquired information unavailable to a clean comparison, even if its final run-time allowance is small.

When some costs are unknown, the practical question is whether they could reverse the reported ranking. Suppose systems A and B meet the same quality and operating requirements at a specified deployment volume. Their reported costs are $\widehat C_A$ and $\widehat C_B$; the omitted costs are $u_A$ and $u_B$. If A's reported saving is $S=\widehat C_B-\widehat C_A>0$, A remains no more expensive than B exactly when
\begin{equation}
u_A-u_B\leq S.
\label{eq:accountingmargin}
\end{equation}
Thus, A's omitted costs can exceed B's by up to the reported saving without reversing the cost ranking. If only intervals $u_j\in[\underline u_j,\overline u_j]$ are available, consider the most unfavorable comparison for A: its largest omitted cost against B's smallest. The ranking holds for every combination in the intervals exactly when $\overline u_A-\underline u_B\leq S$. This identifies which missing cost could change the decision and how large its differential must be. The calculation remains conditional on the stated quality and operating requirements; uncertainty in those requirements needs separate assessment.

Table~\ref{tab:snellenvelope} applies the envelope to the published ICLR 2025 version of \citet{snell2024scaling}. Its purpose is to distinguish the measured comparison from the additional information needed for a deployment decision. The editions of this study differ in their accounting, so the table cites one edition.

\begin{table}[!htbp]
\centering\small
\caption{Resource envelope for one published comparison: the ICLR 2025 edition of \citet{snell2024scaling}, test-time compute versus model scale on MATH. Section and appendix references within the table refer to that paper. ``Not reported'' means the inspected sections do not state the quantity.}
\label{tab:snellenvelope}
\begin{tabular}{@{}L{2.6cm}L{9.6cm}L{2.4cm}@{}}
\toprule
Field & What the study reports & Where\\
\midrule
Task and outcome & MATH, 500-question test split (12k train), PaLM 2-S* base model. Outcome: accuracy of the single returned answer, reported by five difficulty bins defined from the base model's pass@1 over 2,048 samples. Uncertainty intervals: not reported in the inspected sections. & \S3.2, \S4\\[4pt]
Development & A process reward model (PRM) fine-tuned from the base model on Monte Carlo rollout labels (16 samples per question, 16 rollouts per step; no human process labels). A revision model trained by supervised fine-tuning (SFT) on multi-turn data constructed post hoc from 64 parallel samples per question. Compute for either: not reported. Strategy selection uses validation data. & \S3.2, \S5.1, \S6.1; Apps.~F, L\\[4pt]
Run-time allowance & A generation budget of $N$ samples or revisions (up to 256 in search experiments; 128 in the sequential/parallel sweep); beam search capped at 40 expansion rounds; lookahead depth $k$. In \S7, a FLOPs-matched budget at three inference loads, $R=D_{\mathrm{inf}}/D_{\mathrm{pre}}\in\{0.08,0.40,11\}$, where $D_{\mathrm{inf}}$ and $D_{\mathrm{pre}}$ are total inference and pretraining token counts. & \S5.2, \S6.2, \S7\\[4pt]
Run-time consumption & Section~7 uses $6P D_{\mathrm{pre}}$ training FLOPs and $4P D_{\mathrm{inf}}$ inference FLOPs, the latter doubling $2P D_{\mathrm{inf}}$ to include verifier overhead; $P$ denotes parameters. The 2,048 samples per question used to estimate difficulty are explicitly not charged (``our experiments do not account for this cost''). Latency and monetary cost: not modeled. & \S3.2, \S5.2, \S6.1, \S7\\[4pt]
Information access & ``Oracle difficulty'' bins use ground-truth correctness on the test set; ``model-predicted difficulty'' uses the learned verifier on the same 2,048 samples. Strategy choice uses validation data. Few-shot prompts; no retrieval, tools, or environment. & \S3.2, App.~E\\[4pt]
Mechanism & Proposer: few-shot prompted base model, or the fine-tuned revision model. Verifier: last-step PRM for search; a separately trained outcome reward model (ORM) for revisions. Search: best-of-$N$ weighted, beam search, lookahead search. Sequential versus parallel allocation of the generation budget. & \S5.1, \S5.2, \S6; App.~N\\[4pt]
Selection and stopping & Best-of-$N$ weighted (PRM scores summed over candidates sharing a final answer). Beam search stops at end of solution or 40 rounds. Revisions selected by sequential majority vote or verifier, because about 38\% of correct answers become incorrect under naive revision. The evaluated policies do not add an abstention option. & \S5.1, \S6.1; Apps.~C, G, N\\[4pt]
Accounting boundary & FLOPs only. Charged: generation and verifier inference, and for the larger comparison model, pretraining at $6P D_{\mathrm{pre}}$ with data fixed. Explicitly excluded: difficulty estimation. Not itemized: PRM, ORM, and revision-model fine-tuning. & \S3.2, \S5.2, \S6.1, \S7\\[4pt]
Comparison protocol & (i) Compute-optimal strategy versus a best-of-$N$ baseline, reported as up to $4\times$ less compute for equal accuracy. (ii) FLOPs-matched exchange against a model with $14\times$ more parameters (data fixed), by difficulty bin and inference load. One model family, one task. & \S1, \S7\\
\bottomrule
\end{tabular}
\end{table}
\FloatBarrier

Read against the three mismatches of the introduction, the envelope answers each one separately. First, how success was counted. The outcome is the accuracy of the single returned answer, which is the measure the team needs. This mismatch does not arise.

Second, what the system was allowed to know. The headline comparison sorts questions into difficulty bins using the test set's answer key. The study also reports a version that sorts them with its learned verifier. Only that version uses information a deployed system will have, so the answer depends on which version the team reads.

Third, what was counted as cost. Costs are counted in FLOPs only, and two costs are left out. The 2,048 samples per question used to estimate difficulty are explicitly not charged. A deployed system would pay that cost on every question. The training of the process reward model, the outcome reward model, and the revision model is not itemized. A deployment would spread that cost over its volume. The reported saving is the strategy comparison: up to four times less inference compute than a best-of-$N$ baseline at equal accuracy. Whether the saving survives the two omitted costs is the comparison in Eq.~\eqref{eq:accountingmargin}. The study does not report the quantities needed to make it. An unreported quantity is a limit on the present comparison, not evidence that its value is zero or that the study lacked the corresponding procedure.

\subsection{Two legitimate comparison questions}
\label{sec:comparisons}
Controlled comparisons ask what changes when a particular resource or component is varied. They hold the remaining procedure fixed and isolate an intervention. For example, a model-size comparison can retain the same retriever and selection rule, while a retrieval ablation can hold generator and context allowance constant. These designs identify local responses, although they may disadvantage a configuration that needs a different policy.

Deployment comparisons ask which tested system best meets a requirement. They can allow separate optimization of prompts, retrieval, and selection for each model, with comparable tuning opportunities and held-out evaluation. Such a comparison is useful for choosing a system but cannot attribute the observed difference to model size alone. Reporting both views, where feasible, prevents a tension between experimental control and practical optimization from being mistaken for an inconsistency.

Configuration search belongs in the envelope's development field. DSPy and model-selection methods make that search increasingly systematic \citep{khattab2024dspy,chen2025selector}, but systematic search still consumes data and computation. Equal numbers of attempted configurations need not mean equal expense, and unequal tuning can alter the best tested configuration. A report should distinguish total tuning allowance, actual tuning consumption, and the rule used to select the final system.

\subsection{Estimating resource interactions}
\label{sec:interactions}
The envelope records what a study did; estimating how budgets interact is a separate design question. Suppose two budgets have low and high settings and other conditions remain fixed. A basic factorial contrast is
\begin{equation}
\widehat\Delta_{ij}=
\widehat Q(b_i^H,b_j^H)-\widehat Q(b_i^H,b_j^L)
-\widehat Q(b_i^L,b_j^H)+\widehat Q(b_i^L,b_j^L).
\label{eq:interaction}
\end{equation}
It measures interaction over the chosen range on the named score scale. It is not a scale-invariant structural property, nor does it establish effects outside the sampled settings. The same held-out tasks should be used across cells where possible, with repeated stochastic runs and uncertainty that respects task-level dependence. Separate seed variation does not replace variation across tasks.

More levels can estimate curvature and trade-offs at fixed quality. Mechanism changes should be represented by separate curves or factorial factors, and information changes should be labeled explicitly. Where adaptation is central, resource allocation can be evaluated as a policy under a global budget across a query stream, rather than forcing identical per-query expenditure. A policy can then spend more on some tasks while improving average cost, tail behavior, or both. Claims about such gains should report the task mixture on which the policy was learned and the mixture on which it was evaluated.

\section{Research agenda: from local curves to transferable decisions}
\label{sec:agenda}
The next question is whether an allocation rule still holds when the task, the scale, or the operating conditions change. Answering it needs experiments, not another inventory of components. Five priorities follow.

\paragraph{Estimate development versus run-time trade-offs on a common outcome.}
Jointly vary model size, training duration, and inference policy. Measure returned-answer accuracy and actual cost, and report coverage where relevant. Fit on a restricted range and test predictions on withheld scales and deployment volumes. A training--inference law is more useful when it predicts the ranking of feasible allocations beyond the points used to fit it. Failures of transfer should be related to changed data regimes, output lengths, or selection quality rather than absorbed into a universal exponent.

\paragraph{Separate information acquisition from information access.}
For retrieval, cross model scale with corpus size and corpus overlap. Separately vary retriever quality and processing budget. Include reused training data, new but distribution-matched data, and genuinely new information relevant to the task. For causal tasks, distinguish additional observational samples from interventions or justified identifying assumptions. This design tests whether the return comes from more information, a better route to existing information, or more computation over the same evidence.

\paragraph{Measure verifier behavior under the search distribution.}
Cross generator and verifier quality under a common total allowance. Report coverage, conditional selection accuracy, abstention, and wrong-return probability. Evaluate verifier calibration on the candidates produced by large or adaptive searches, not only on a fixed independent test set. Vary correlation among proposals and track how the marginal value of sampling changes. A resulting allocation rule should predict when spending on selection is more effective than spending on additional proposals, including the development cost of each option.

\paragraph{Distinguish more interaction from better interaction.}
Cross an agent's allowed horizon with its training regime and per-step deliberation budget. Control tool access and environment versions. Record observations and resets. Compare serial and parallel coordination under both work and latency constraints. Evaluate whether improvements persist when the task becomes longer or the environment changes. This can distinguish a general ability to use additional observations from exploitation of one benchmark's feedback or the option to reset the environment.

\paragraph{Test allocation policies under workload uncertainty.}
Resource decisions depend on demand volume, query difficulty, failure costs, and prices. Evaluate policies under shifts in these quantities. Report regret or excess cost relative to an explicitly defined feasible comparator. Include tail constraints and the expense of adapting a policy. A rule that works only after extensive task-specific tuning may still be useful, but its tuning burden should be part of the claimed advantage.

These priorities provide falsifiable comparisons without presuming their outcomes. Retrieval may cease to be cost-effective after serving costs are charged. Stronger verification may fail when candidates shift. Longer interaction may add unproductive actions. A small model's inference advantage may disappear on tasks where useful proposals are too rare.

\section{Limitations and conclusion}
\label{sec:conclusion}
This review has four principal limitations. Its purposive search cannot establish exhaustive coverage, and the rapid emergence and revision of preprints make the synthesis time-bounded. Its empirical center is language-model systems, so extension to robotics, perception, or other domains requires additional evidence. The reviewed studies use heterogeneous tasks, model access, and accounting conventions, preventing a justified pooled scaling coefficient. Finally, the analytical examples isolate mechanisms through simplifying assumptions; they establish possibilities and reporting needs, not estimated effects or optimal deployment policies.

The capability surface likewise does not solve the optimization problem it describes. It is high-dimensional and expensive to estimate, information may be only partially observable, and mechanisms can change during development. A useful empirical program will require selective experimental designs, uncertainty-aware models, and careful task analysis. The proposed envelope organizes these requirements, but its utility and reporting burden should themselves be evaluated in practice.

The central conclusion is that scaling evidence becomes more actionable when the unit of analysis is the complete procedure producing a scored outcome. Pretraining, inference, retrieval, verification, and interaction studies establish important conditional results. Once quality, access, and costs are aligned, the relevant question becomes which feasible resource allocation reaches a task requirement most effectively.

For the team in the introduction, the answer to this review's question is a test with three parts. A reported result justifies a different allocation when its score was counted on the returned answer, when the system was allowed to know no more than the team's system will know, and when the cost that produced it was counted at the team's boundary. When any of the three is missing, the envelope says which one, and that is the experiment to run next.

\appendix
\section{Derivations and reproducibility of the analytical examples}
\label{app:derivations}
\subsection{Increasing transformations of the score}
Applying the chain rule twice gives Eq.~\eqref{eq:metric}. With $g=b_1b_2$ and $h(z)=-1/z$, $h'(z)=z^{-2}>0$ while direct differentiation gives $\partial^2[-(b_1b_2)^{-1}]/\partial b_1\partial b_2=-1/(b_1^2b_2^2)$. Along a regular level set, $dg=g_i\,db_i+g_j\,db_j=0$. Multiplication by positive $h'(g)$ leaves its slope unchanged. This elementary argument separates numerical interaction signs from ordinally invariant trade-offs; it is not claimed as a new mathematical result.

\subsection{Coverage and selection}
Under independent candidates of common success probability $p$, the probability all $n$ fail is $(1-p)^n$. When selection returns a member of the candidate set, a correct return requires a set containing a correct member, so the conditional-probability rule gives Eq.~\eqref{eq:selection}. For the example, feasible counts are $\lfloor12/(1+0.1)\rfloor=10$ and $\lfloor12/(1+0.8)\rfloor=6$. Multiplying the associated coverage by the assumed conditional selection accuracy yields $0.5802068$ and $0.7009632$, respectively.

\subsection{First-accept stopping}
A correct acceptance at attempt $t$ requires $t-1$ nonacceptances and a correct acceptance, giving $(1-a)^{t-1}ps$. Summing the geometric series to $n$ gives Eq.~\eqref{eq:stopping}; substituting $(1-p)f$ gives the wrong-return probability. No acceptance has probability $(1-a)^n$. The expected number of attempts is the sum of the probabilities of reaching each attempt, $\sum_{t=1}^n(1-a)^{t-1}$. If $a=0$, no return occurs and all $n$ attempts are made.

\subsection{Observational equivalence and estimation error}
Both causal models have zero means, unit marginal variances, and covariance $\rho$, and therefore the same joint Gaussian observational law. Let $Z$ be any estimate of the intervention mean based only on that law's samples and common auxiliary information. Its distribution is the same in both worlds. Writing $d=\rho x$,
\begin{equation}
\frac{\E[(Z-d)^2]+\E[Z^2]}{2}
=\E[(Z-d/2)^2]+\frac{d^2}{4}\geq\frac{d^2}{4}.
\end{equation}
The larger of the two risks is at least their average. This is a two-world illustration of a known identification issue; supplying distinguishing information changes the premise.

\section*{Statements and declarations}
\paragraph{Funding.} The author received no funding for this work.
\paragraph{Competing interests.} The author declares no competing interests.
\paragraph{Data and code availability.} No new empirical dataset was generated or analyzed. All numerical illustrations are based on assumptions stated in the text and can be reproduced from the formulas given.
\paragraph{Use of generative AI.} Generative AI tools (ChatGPT, Claude) assisted with literature search, \LaTeX{} preparation, and the code for the analytical illustrations. The author read and verified the manuscript, independently checked the analytical examples, and is responsible for its content. Generative AI is not an author.

\bibliographystyle{abbrvnat}
\bibliography{references_AIR}
\end{document}